\documentclass{article}

\usepackage[main, final]{neurips_2026}

\usepackage[utf8]{inputenc} 
\usepackage[T1]{fontenc}    
\usepackage{hyperref}       
\usepackage{url}            
\usepackage{booktabs}       
\usepackage{amsfonts}       
\usepackage{nicefrac}       
\usepackage{microtype}      
\usepackage{xcolor}         
\usepackage{graphicx}

\usepackage{listings}
\usepackage{multirow}
\usepackage{amsmath}

\lstdefinelanguage{json}{
    basicstyle=\ttfamily\footnotesize,
    breaklines=true,
    frame=single,
    showstringspaces=false,
    columns=fullflexible,
    string=[s]{"}{"},
    stringstyle=\color{teal!70!black},
    comment=[l]{//},
    commentstyle=\color{gray},
}

\title{Meta-Agent: From Task Descriptions to Verified Multi-Agent Systems}

\author{
  Andy Xu \\
  Dartmouth College\\
  \texttt{andy.xu.26@dartmouth.edu}
  \and 
  Yu-Wing Tai\\
  Dartmouth College\\
  \texttt{yu-wing.tai@dartmouth.edu}}

\begin{document}

\maketitle

\begin{abstract}
AI agents are increasingly used to solve complex, multi-step tasks, but existing multi-agent frameworks remain brittle as workflows grow in scale and depth. Small errors at intermediate stages can propagate through agent interactions, while insufficient grounding and weak verification mechanisms further limit reliability. We present \textbf{Meta-Agent}, a two-phase framework that automatically constructs and executes specialized multi-agent systems from natural-language task descriptions. In the construction phase, a task planner decomposes a problem into a directed acyclic graph of agent specifications with explicit input/output contracts and verification criteria. A web search module grounds each specification with external evidence, and a code generation module produces system prompts and tool configurations. A construction-time verification stage then validates generated artifacts and triggers targeted regeneration when failures are detected. In the execution phase, a coordinator dispatches subtasks across the agent graph while execution-time verification gates intermediate outputs. We further introduce a three-level error attribution mechanism that distinguishes local, upstream, and structural failures, enabling targeted recovery strategies ranging from localized retries to partial re-execution and re-decomposition. We evaluate Meta-Agent across coding, contextual learning, and open-ended reasoning tasks. Experiments against strong multi-agent baselines and ablation studies demonstrate consistent improvements in task success rate, error recovery, and workflow stability. The results highlight the importance of tightly integrating planning, grounding, and verification for building reliable multi-agent systems.
\end{abstract}
\section{Introduction}

\begin{figure}[t]
\centering
\includegraphics[width=\linewidth]{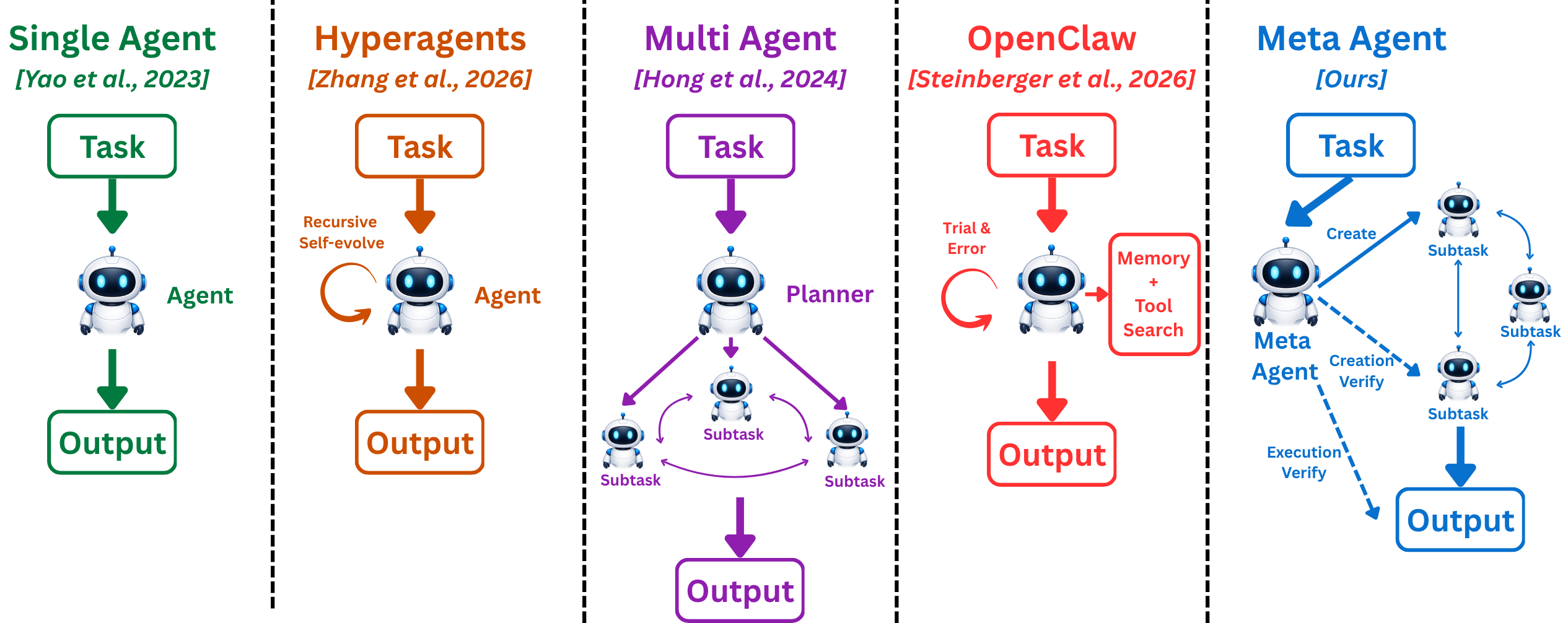}\\
\vspace{-0.1in}
\caption{\textbf{Comparison of agent paradigms.} We contrast five approaches to constructing AI agents for solving tasks. \textbf{Single Agent}~\citep{yao2023react} uses one LLM that reasons and acts in a loop until completion. \textbf{Hyperagent}~\citep{zhang2026hyperagents} augments a single agent with recursive self-evolution, iteratively refining its behavior across turns. \textbf{Multi Agent}~\citep{hong2024metagpt} decomposes tasks via a planner that assigns subtasks to specialized workers, which coordinate and merge outputs. \textbf{OpenClaw}~\citep{openclaw} is a personal assistant that learns online from interaction signals (user replies, tool outputs), with persistent memory and tool search. \textbf{Meta Agent (ours)} introduces a meta-level agent that constructs a task-specific multi-agent system at inference time: it both \emph{creates} worker agents (solid arrows) and \emph{verifies} them (dashed arrows). Unlike prior paradigms, it treats the agent system itself as a synthesize-and-validate artifact per task.}
\vspace{-0.2in}
\label{fig:comparison}
\end{figure}

Real-world problems are inherently complex, underspecified, and multi-stage. Users often provide only partial or ambiguous task descriptions, leaving critical details, such as decomposition strategies, required knowledge sources, and validation criteria, unstated. Solving such tasks therefore requires not only executing instructions, but \emph{inferring how to structure the problem itself}. This challenge is amplified in long-horizon settings, where errors made early in the process may remain undetected and propagate through subsequent steps, ultimately compromising the final outcome.

Recent advances in large language model (LLM) agents have enabled systems that can plan, use tools, and execute multi-step workflows. Benchmarks such as AgentBench~\cite{liu2023agentbench} highlight that performance in these settings depends critically on decision-making and coordination rather than raw language modeling ability alone. To address increasing task complexity, a growing line of work explores multi-agent systems, where specialized agents collaborate to solve subtasks~\cite{hong2024metagpt, chen2025aflow}. However, existing approaches typically rely on fixed or manually designed agent structures, which assume that the decomposition of the task is known a priori. In practice, this assumption rarely holds: the optimal structure of a multi-agent system depends on the task itself, which must be inferred from incomplete user input.

This motivates a shift from designing \emph{agents} to designing \emph{systems of agents}. In this work, we introduce the concept of a \textbf{meta-agent}: an agent whose role is to construct other agents tailored to a given task. Rather than directly solving the problem, the meta-agent interprets a natural-language task description, decomposes it into a structured workflow, and generates a collection of specialized agents with well-defined roles, interfaces, and dependencies. This perspective reframes multi-agent systems as \emph{generated artifacts} rather than fixed architectures, enabling adaptation to diverse and previously unseen tasks. Figure~\ref{fig:comparison} illustrates how this paradigm differs from prior approaches.

A central challenge in this paradigm is reliability. Automatically generated agent systems are particularly susceptible to cascading failures: errors in task decomposition, missing or incorrect grounding, and poorly specified intermediate outputs can silently propagate across agents. Existing approaches often rely on implicit self-reflection or post hoc correction, which are insufficient for long-horizon workflows where failures compound over time. We argue that reliable meta-agent systems require \emph{explicit, structured verification} that is tightly integrated throughout both the construction and execution processes.

To this end, we present \textbf{Meta-Agent}, a two-phase framework for transforming task descriptions into \emph{verified} multi-agent systems. In the \emph{construction phase}, the meta-agent decomposes the task into a directed acyclic graph (DAG) of agent specifications, each annotated with input/output contracts and verification criteria. Crucially, verification is introduced at construction time: generated agent specifications, prompts, and tool configurations are validated against these criteria, and failures trigger targeted regeneration of specific components rather than global retries. This process ensures that the resulting agent system is structurally sound before execution begins.

In the \emph{execution phase}, verification continues to play a central role. Each agent's output is gated by task-specific validation checks, preventing erroneous intermediate results from propagating downstream. We further introduce a three-level error attribution mechanism that distinguishes \emph{local} failures (within an agent), \emph{upstream} failures (inherited from prior agents), and \emph{structural} failures (arising from flawed task decomposition). This attribution enables targeted recovery strategies, including localized retries, partial re-execution, and, when necessary, re-construction of the agent graph. Together, these mechanisms form a closed-loop system in which planning, grounding, execution, and verification are tightly coupled.

We evaluate Meta-Agent across diverse domains, including coding, contextual learning, and open-ended reasoning tasks. Compared to strong multi-agent baselines, our approach achieves consistent improvements in task success rate, robustness to intermediate errors, and stability in long-horizon workflows. Ablation studies demonstrate that the integration of verification, particularly during the construction phase, is critical to these gains.

In summary, this work makes three main contributions: (1) we introduce the meta-agent paradigm for automatically generating task-specific multi-agent systems from natural-language descriptions; (2) we propose a unified framework that integrates planning, grounding, and \emph{verification} across both construction and execution; and (3) we demonstrate that explicit verification loops and error attribution mechanisms are key to scaling agent systems reliably in complex, real-world settings.
\section{Related Work}
\label{sec:related}

\begin{table}[t]
\centering
\footnotesize
\setlength{\tabcolsep}{3pt}
\renewcommand{\arraystretch}{1.2}

\begin{tabular}{lccc}
\toprule
\textbf{Paradigm} 
& \textbf{Structure} 
& \textbf{Agent Construction} 
& \textbf{Verification} \\
\midrule

ReAct~\cite{yao2023react} 
& Fixed loop 
& Fixed policy 
& $\times$ \\

HyperAgent~\cite{zhang2026hyperagents} 
& Single agent 
& Self-evolving policy 
& $\times$ \\

MetaGPT~\cite{hong2024metagpt} 
& Predefined workflow 
& Plan decomposition 
& Local \\

AutoGen~\cite{wu2023autogen} 
& Conversational graph 
& Runtime orchestration 
& Post-hoc \\

OpenClaw~\cite{openclaw} 
& Lifelong agent 
& Adaptive agent 
& $\times$ \\

\textbf{Meta-Agent (ours)} 
& \textbf{Generated system} 
& \textbf{System synthesis} 
& \textbf{Construction + execution} \\
\bottomrule
\end{tabular}

\caption{\textbf{Comparison of agent paradigms.} Prior work operates at different levels of agent design, ranging from fixed reasoning loops (ReAct), self-evolving single agents (HyperAgent), predefined multi-agent workflows (MetaGPT, AutoGen), and lifelong adaptive agents (OpenClaw). In contrast, Meta-Agent constructs a fully executable multi-agent system from task descriptions, explicitly synthesizing agent structure and enforcing unified verification across both construction and execution stages.}
\vspace{-0.2in}
\label{tab:agent_paradigms}
\end{table}

\subsection{Agent Paradigms and Multi-Agent Workflow Generation}

Recent advances in large language model (LLM) agents span a spectrum of paradigms, differing in how agent behavior is defined: from fixed reasoning loops, to self-evolving policies, to multi-agent coordination, and ultimately to persistent or automatically synthesized systems. We summarize this spectrum in Table~\ref{tab:agent_paradigms}.

At the simplest level, single-agent frameworks such as ReAct~\cite{yao2023react} interleave reasoning and action in a unified loop, treating the agent as a monolithic policy. A growing line of work studies \emph{self-improving agents}, where the policy is recursively refined. HyperAgent-style systems~\cite{zhang2026hyperagents} enable agents to iteratively modify their own reasoning and improvement mechanisms, connecting to earlier ideas in recursive self-improvement such as Gödel Machines~\cite{zhang2025dgm}, self-referential learning~\cite{kirsch2022eliminating, irie2022modern}, and recent self-improving coding agents and recursive language models~\cite{robeyns2025sica, zhang2025recursive}. This direction is further influenced by open-ended and evolutionary paradigms such as AI-GAs~\cite{clune2019aigas} and novelty search~\cite{lehman2011evolving}.

Multi-agent systems instead decompose tasks into specialized roles. MetaGPT~\cite{hong2024metagpt} adopts predefined workflows, while AutoGen~\cite{wu2023autogen} and CAMEL~\cite{li2023camel} enable flexible agent coordination. GPT-Swarm~\cite{zhuge2024gpt} further models agent systems as optimizable graphs, allowing structural evolution.

Recent work increasingly automates agent design. ADAS~\cite{hu2025adas} and Agentic Context Engineering~\cite{zhang2025ace} jointly optimize prompts, tools, and workflows, while PromptBreeder~\cite{fernando2023promptbreeder} and Self-Taught Optimizer~\cite{zelikman2024stop} evolve optimization strategies. Systems such as OS-Copilot~\cite{wu2024oscopilot} and Voyager~\cite{wang2024voyager} further demonstrate self-improvement through tool use and interaction. Persistent and adaptive agents extend this paradigm. OpenClaw~\cite{openclaw} enables lifelong learning from interaction, while Evo-Memory~\cite{wei2025evomemory} and MemSkill~\cite{zhang2026memskill} evolve memory structures over time.
Finally, large-scale agentic systems such as the AI Scientist~\cite{lu2024aiscientist}, SWE-agent-style systems~\cite{jimenez2024swebench}, and AlphaEvolve~\cite{novikov2025alphaevolve} demonstrate end-to-end automation of research, coding, and algorithm discovery.

Despite these advances, prior work operates at different abstraction levels, either fixed agents, self-modifying agents, or optimized multi-agent systems. In contrast, Meta-Agent synthesizes the entire agentic system, including structure, roles, dependencies, and verification, from task descriptions.

\subsection{Verification and Reliable Agent Systems}

As agent systems grow in complexity, ensuring reliability across long-horizon workflows has become a central challenge. Early approaches rely on self-reflection or heuristic retry mechanisms, which are often insufficient for preventing cascading errors. Recent works explore self-correction and critique-based refinement, where LLMs iteratively improve their outputs using feedback signals or tool-based critiques~\cite{chen2023selfdebug, gou2023critic}. However, such approaches lack formal guarantees and may still propagate hidden inconsistencies.

Recent work integrates verification more explicitly into agent pipelines. VeriMAP~\cite{xu2025verimap} embeds verification functions directly into multi-agent planning graphs, enabling runtime validation of intermediate outputs. VeriPlan~\cite{lee2025veriplan} introduces formal verification of LLM-generated plans by mapping constraints into temporal logic specifications and checking correctness against them. More broadly, formal methods from classical planning and verification, such as temporal logic and model checking~\cite{baier2008model, dixon2014temporal}, have been adapted to ensure correctness in autonomous systems.

In parallel, several works propose learning-based or execution-based verification strategies. For example, execution-guided verification has been explored in code and reasoning tasks, where outputs are validated against runtime behavior or external tools~\cite{ni2023lever}. Step-aware verification further improves reasoning correctness by validating intermediate steps in multi-hop inference~\cite{li2023stepaware}. Collaborative or multi-agent verification has also been studied, where multiple agents cross-check each other's outputs to improve robustness~\cite{liang2024collaborative}. 

System-level reliability has also been studied through infrastructure-oriented designs. AgentGit~\cite{agentgit2025} introduces version control mechanisms for agent workflows, enabling branching, rollback, and systematic comparison of execution trajectories. These approaches improve reproducibility and debugging of multi-agent systems.

However, existing methods primarily verify fixed or already-instantiated agent structures. Meta-Agent differs by integrating verification at construction time, ensuring that agent specifications, tool configurations, and inter-agent dependencies are validated before execution, and enabling targeted reconstruction when failures occur.

\subsection{Contextual Learning and Reasoning Benchmarks}

Evaluating agent systems requires benchmarks that measure adaptation to contextual information and long-horizon reasoning. CL-Bench~\cite{clbench2026} targets contextual learning across diverse settings. General reasoning benchmarks (MMLU~\cite{hendrycks2021mmlu}, BIG-Bench Hard~\cite{bbh2022}, GPQA~\cite{gpqa}) are increasingly saturated, while mathematical benchmarks such as GSM8K~\cite{gsm8k}, MATH~\cite{math_benchmark}, and Omni-MATH~\cite{omnimath2024} stress multi-step symbolic reasoning. Agent-oriented benchmarks such as AgentBench~\cite{liu2023agentbench} evaluate tool use and long-horizon decision-making, exposing the fragility of current systems under extended execution. These benchmarks largely test reasoning or tool use in isolation; Meta-Agent instead targets system-level correctness under explicit verification constraints.
\section{Method}

\begin{figure}[t]
  \centering
  \includegraphics[width=0.7\linewidth]{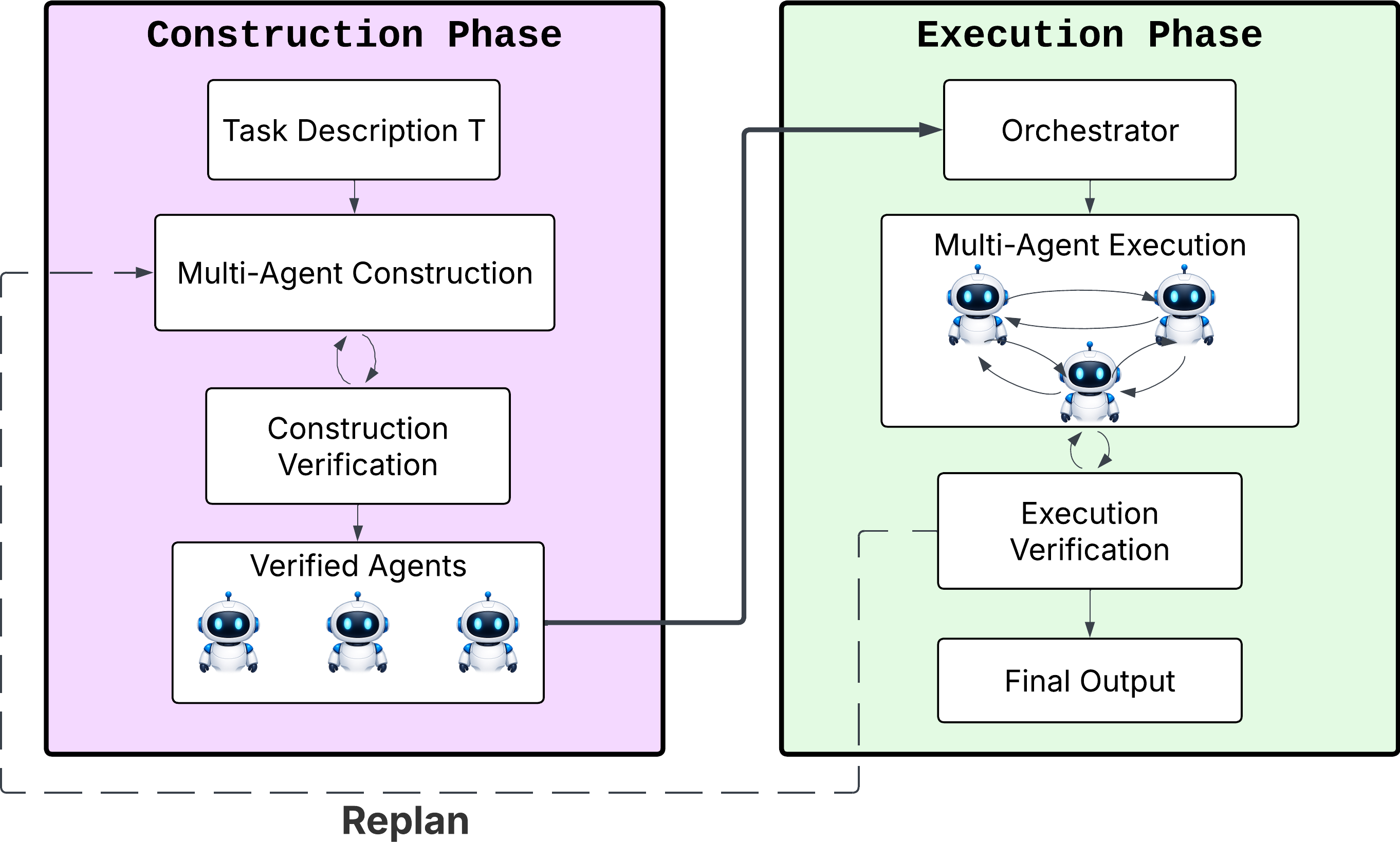}\\
  \vspace{-0.1in}
  \caption{\textbf{Overview of the Meta-Agent.} Phase~1
  (left) transforms a task description into a DAG of verified agents
  through a pipeline consisting of prompt analysis, architecture planning, API research, code generation, and verification. Verification failures are typed and routed back to the responsible upstream stage. Phase~2 (right) executes the resulting multi-agent DAG under continuous verification. A coordinator dispatches agents, an in-memory context store routes intermediate outputs, and typed error attribution distinguishes local, upstream, and structural failures to determine the corrective action.}
  \vspace{-0.15in}
  \label{fig:architecture}
\end{figure}

\subsection{Overall Framework}

We propose \textbf{Meta-Agent}, a two-phase framework that transforms a natural-language task description into a verified multi-agent system and executes it under continuous validation. Given a task description $T$, the goal is to construct a set of specialized agents $\{a_1, \dots, a_n\}$ together with a directed acyclic graph (DAG) $G = (V, E)$ that defines their dependencies. Each agent is associated with an explicit input/output contract and a set of verification criteria, so that the resulting system is not only executable but also verifiable at every stage.

As illustrated in Figure~\ref{fig:architecture}, the framework consists of a \emph{construction phase} followed by an \emph{execution phase}. In the construction phase, the system compiles the task description $T$ into a complete specification of a multi-agent system. This process begins by parsing $T$ into a structured representation that captures the task goal, constraints, and relevant context. Based on this representation, a planner decomposes the task into a small number of subtasks and organizes them into a DAG. For each node in the graph, the system produces an agent specification that includes a role description, a tool interface, an input/output schema, and a set of verification criteria.

To ensure that each agent is grounded in external knowledge, the system augments these specifications with information retrieved from external sources, such as APIs or documents. The final step of the construction phase compiles each specification into an executable agent with a standardized interface. Importantly, the output of this phase is a fully specified and executable multi-agent system, rather than a high-level plan.

In the execution phase, a coordinator executes the agent graph according to the DAG structure. Agents are dispatched when their dependencies are satisfied, and their outputs are stored and passed to downstream agents. Unlike conventional pipelines, execution is tightly coupled with verification: every intermediate output is validated before it is consumed. This design prevents errors from propagating across agents and enables robust execution in long-horizon tasks.

A central design principle of Meta-Agent is that verification is treated as a first-class component of the system. Rather than applying validation only after execution, the framework integrates verification throughout both construction and execution. We formalize this mechanism as a unified verification loop in the next subsection.

\subsection{Verification Loop and Error Attribution}

The core contribution of Meta-Agent is a unified verification loop that detects, attributes, and corrects errors throughout the system lifecycle. Instead of a purely feed-forward pipeline, the framework operates as a closed loop:
\begin{equation}
\text{Generate} \;\rightarrow\; \text{Verify} \;\rightarrow\; \text{Attribute} \;\rightarrow\; \text{Refine}.
\end{equation}
This loop is applied both during agent construction and during execution, ensuring that errors are identified early and corrected with minimal cost.

\noindent\textbf{Construction-time verification.}
During the construction phase, each generated agent is validated before being admitted into the system. Let $\sigma_i$ denote the specification of agent $a_i$, and let $\hat{a}_i$ denote its generated implementation. The goal of construction-time verification is to ensure that $\hat{a}_i$ satisfies the specification $\sigma_i$. This is achieved through two complementary checks.

First, a static verification step ensures that the generated implementation is well-formed. This includes validating the code structure, checking that required interfaces are present, and confirming that the agent can be instantiated without runtime errors. Second, a behavioral verification step evaluates whether the agent behaves as intended. A verifier model simulates the execution of $\hat{a}_i$ on representative inputs and checks whether the outputs satisfy the declared input/output contract and behavioral assertions.

Crucially, verification does not produce a binary decision alone. Instead, it returns a typed failure signal that identifies the source of the error. We define a failure type $f \in \mathcal{F}$, where $\mathcal{F}$ includes categories such as specification mismatch, grounding failure, and contract violation. Each failure type is associated with a specific upstream stage. For example, a grounding failure triggers a re-execution of the knowledge retrieval stage, while a contract violation triggers a revision of the task decomposition. This typed routing allows the system to correct errors locally without recomputing the entire pipeline.

\noindent\textbf{Execution-time verification.}
Verification continues during execution to ensure that intermediate results remain valid. Let $y_i$ denote the output of agent $a_i$, and let $\mathcal{C}_i$ denote its verification criteria. Before $y_i$ is passed to downstream agents, a verifier checks whether
\begin{equation}
y_i \in \mathcal{C}_i,
\end{equation}
where $\mathcal{C}_i$ encodes schema constraints, behavioral assertions, and forbidden patterns. If the output fails verification, it is not propagated further, and the system initiates a recovery process.

\noindent\textbf{Typed error attribution.}
To enable efficient recovery, Meta-Agent introduces a three-level error attribution mechanism. Given a failure at agent $a_i$, the system classifies the error into one of three categories.

A \emph{local error} occurs when $a_i$ produces an incorrect output despite receiving correct inputs. In this case, the system retries the same agent with additional feedback from the verifier. An \emph{upstream error} occurs when the failure originates from one of the dependencies of $a_i$. The system identifies the responsible upstream agent and re-executes it before retrying $a_i$. A \emph{structural error} indicates that the task decomposition itself is flawed, for example due to an incorrect input/output contract. In this case, the system escalates to the construction phase and reconstructs the affected part of the agent graph.

This attribution mechanism ensures that the cost of recovery scales with the locality of the error. Most failures can be resolved through local retries, while more expensive operations such as re-planning are invoked only when necessary. This design prevents cascading failures and significantly improves robustness in long-horizon workflows.

\subsection{Running Example}
We illustrate Meta-Agent end-to-end on a function-completion task (full
pipeline trace and additional running examples in Appendix~\ref{app:humaneval-trace}, ~\ref{app:math-trace} and ~\ref{app:drop-trace}).

\noindent\textbf{Construction phase.}
Given a task description $T$ asking the system to complete Python
functions from a signature and docstring, the Meta-Agent compiles $T$
into a structured intent (goal, domain, constraints, task-category
examples retrieved via web search). The swarm planner decomposes the
task into four specialist agents in a DAG: (1) a \emph{specification
analyst} that parses signature and docstring into parameter types,
behavioral rules, edge cases, and resolved ambiguities; (2) a
\emph{strategy planner} that selects an algorithm, data structures,
and edge-case handling plan; (3) a \emph{code synthesizer} that emits
the implementation as a single fenced block; and (4) a \emph{code
verifier} that audits the output and emits a \texttt{PASS}/\texttt{FAIL}
verdict. Each agent has an explicit I/O contract: the synthesizer's,
for instance, requires the exact input signature, verbatim docstring,
and a guard clause for every planned edge case. Construction-time
verification catches adherence failures and triggers regeneration
before execution.

\noindent\textbf{Execution phase.}
At runtime, the coordinator runs the four agents in topological order.
For the function:
\begin{verbatim}
def has_close_elements(numbers: List[float],
                       threshold: float) -> bool:
    """Check if any two numbers in the list
    are closer than the given threshold."""
\end{verbatim}
the analyst extracts implied edge cases (empty list, single element,
zero threshold) and the rule \emph{``return True if any pair distance
$<$ threshold''}. The planner selects nested pairwise comparison with
early returns; the synthesizer produces the implementation; the verifier
confirms signature match, verbatim docstring, edge-case coverage, and
no third-party imports. The verified code runs against hidden unit
tests in a sandbox.

\noindent\textbf{Error attribution and recovery.} Failures are localized via the three-way classification. If the
synthesizer uses $\leq$ instead of $<$ because the analyst never flagged
the strict inequality, the error is upstream and the analyst is re-run
with feedback; if the analyst flagged it but the synthesizer ignored
it, the synthesizer retries locally; if the planner selected an
incompatible algorithm (e.g., sorting that destroys pair ordering),
the failure is a contract violation and the planner is re-invoked;
and if no existing agent could plausibly resolve the issue, the failure
is structural and the coordinator re-plans the relevant subgraph.
This example shows how Meta-Agent unifies planning, decomposition,
execution, and verification in one pipeline, with construction-time
verification preventing cascading failures.

\subsection{Design Choices and Implementation Details}

We highlight a small number of key design choices that are critical to the effectiveness of Meta-Agent. These choices focus on improving reliability and enabling efficient error recovery in multi-agent systems.

\noindent\textbf{Design-time grounding.}
We perform knowledge grounding during the construction phase rather than at execution time. External information, such as APIs and reference documents, is retrieved before agent generation and incorporated into the agent specification. This ensures that each agent is built with access to the required knowledge, reducing ambiguity during execution and lowering the likelihood of verification failures caused by missing or incorrect information.

\noindent\textbf{Verification-driven refinement.}
Verification is tightly integrated into the generation process. When a failure is detected, the verifier produces structured feedback that is used to guide the next generation attempt. This creates a refinement loop in which agents are iteratively improved until they satisfy their specifications. In practice, this significantly reduces misalignment between planned behavior and actual implementation.

\noindent\textbf{Typed error attribution for localized recovery.}
We adopt a structured error attribution mechanism that classifies failures into local, upstream, and structural categories (Section~3.2). This allows the system to apply targeted corrections, such as retrying a single agent or re-executing a dependency, instead of restarting the entire pipeline. As a result, the cost of recovery scales with the scope of the error.

\noindent\textbf{Trade-off between verification cost and robustness.}
Verification introduces additional computational overhead, since each intermediate result must be validated before propagation. However, this cost is offset by improved robustness: early detection of errors prevents cascading failures in long-horizon workflows. Empirically, we find that this trade-off leads to more stable execution and higher task success rates.

These design choices work together to ensure that Meta-Agent remains both reliable and efficient when applied to complex, multi-step tasks.
\section{Experiments}
\label{sec:experiments}

\subsection{Experimental Setup}

\noindent\textbf{Benchmarks.}
We evaluate Meta-Agent on six benchmarks across three categories,
following the protocol of AFlow~\cite{zhang2024aflow} for benchmark
selection, sample sizes, and metrics. For code generation, we use
HumanEval~\cite{chen2021codex} (pass@1) and
MBPP~\cite{austin2021mbpp} (pass@1). For
mathematical reasoning, we use GSM8K~\cite{cobbe2021gsm8k} and MATH~\cite{hendrycks2021math}
(Combinatorics \& Probability,
Number Theory, Pre-algebra, and Pre-calculus, solve rate). For reading
comprehension, we use HotpotQA~\cite{yang2018hotpotqa} and
DROP~\cite{dua2019drop}.

\noindent\textbf{Implementation.}
Meta-Agent's components (planner, code generation module, verifier,
and per-agent executors) can be instantiated with different language
models. We use GPT-4o-mini for the main comparison
(Section~\ref{sec:main_results}) to match AFlow's executor, and
additionally report Claude Sonnet~4.6 in
Section~\ref{sec:exec_agnostic}. Verification is applied at both
construction and execution time. The DAG structure, agent
specifications, and tool configurations are generated per task; no
templates are reused across benchmarks. We report single-run scores.

\noindent\textbf{Baselines.}
We compare against the seven baselines from AFlow's main
table~\cite{zhang2024aflow}, all evaluated on the same splits with
GPT-4o-mini: single-pass prompting (IO), structured prompting
(CoT~\cite{wei2022cot}, CoT-SC~\cite{wang2022selfconsistency},
MedPrompt~\cite{nori2023medprompt},
MultiPersona~\cite{wang2024multipersona}), self-correction
(Self-Refine~\cite{madaan2023selfrefine}), and automated agent design
(ADAS~\cite{hu2024adas} and AFlow). Baseline numbers are taken
from~\cite{zhang2024aflow}.

\subsection{Main Results}
\label{sec:main_results}

Table~\ref{tab:main_results} reports Meta-Agent's performance against
the AFlow baseline suite under a matched GPT-4o-mini executor.
Meta-Agent achieves the highest score on five of six benchmarks,
trailing AFlow only on HotpotQA. The average of 82.7 improves on
AFlow (80.3) by 2.4 points and on the strongest non-agentic baseline,
CoT-SC (76.0), by 6.7 points.

\begin{table}[t]
\centering
\small
\setlength{\tabcolsep}{4pt}
\begin{tabular}{lccccccc}
\toprule
\textbf{Method} & \textbf{DROP} & \textbf{HumanEval} & \textbf{HotpotQA} & \textbf{MBPP} & \textbf{GSM8K} & \textbf{MATH} & \textbf{Avg.} \\
\midrule
IO                                                & 68.3 & 87.0 & 68.1 & 71.8 & 92.7 & 48.6 & 72.8 \\
CoT~\cite{wei2022cot}                             & 78.5 & 88.6 & 67.9 & 71.8 & 92.4 & 48.8 & 74.7 \\
CoT-SC~\cite{wang2022selfconsistency}             & 78.8 & 91.6 & 68.9 & 73.6 & 92.7 & 50.4 & 76.0 \\
MedPrompt~\cite{nori2023medprompt}                & 78.0 & 91.6 & 68.3 & 73.6 & 90.0 & 50.0 & 75.3 \\
MultiPersona~\cite{wang2024multipersona}          & 74.4 & 89.3 & 69.2 & 73.6 & 92.8 & 50.8 & 75.1 \\
Self-Refine~\cite{madaan2023selfrefine}           & 70.2 & 87.8 & 60.8 & 69.8 & 89.6 & 46.1 & 70.7 \\
ADAS~\cite{hu2024adas}                            & 76.6 & 82.4 & 64.5 & 53.4 & 90.8 & 35.4 & 67.2 \\
AFlow~\cite{zhang2024aflow}                       & 80.6 & 94.7 & \textbf{73.5} & 83.4 & 93.5 & 56.2 & 80.3 \\
\midrule
\textbf{Meta-Agent (ours)} & \textbf{82.7} & \textbf{96.0} & 69.5 & \textbf{84.6} & \textbf{93.7} & \textbf{69.6} & \textbf{82.7} \\
\bottomrule
\end{tabular}
\caption{Comparison against baselines on six benchmarks.
All methods use GPT-4o-mini and the test splits established
by~\cite{zhang2024aflow}. Bold marks the best score in each column.}
\label{tab:main_results}
\vspace{-0.2in}
\end{table}

The 2.4-point average improvement is driven by reasoning-heavy
benchmarks. The largest gain over AFlow is on MATH (+13.4 points on a
benchmark restricted to level-5 competition problems), which we
attribute to the planner's ability to allocate dedicated sub-agents
for symbolic manipulation, intermediate verification, and answer
extraction. DROP shows a similar pattern (+2.1 over AFlow, +3.9 over
CoT-SC), consistent with the observation that constructed agent
decompositions help most when the task admits a multi-stage solution
structure. Code-generation gains are smaller (HumanEval +1.3, MBPP
+1.2 over AFlow), partly reflecting ceiling effects under
GPT-4o-mini; the gap to non-agentic baselines is much wider than the
gap to AFlow, indicating that most of the benefit on code tasks comes
from agentic structure itself. GSM8K is largely saturated, with all
methods within 4.1 points of each other.

\subsection{Ablation Study}
\label{sec:ablation}

To isolate the contribution of each construction-phase component, we
ablate them one at a time on DROP under GPT-4o-mini
(Table~\ref{tab:ablation}). Every component contributes positively.
The two largest drops come from removing verification ($-7.1$) and
API research ($-5.5$), supporting our central claim that
construction-time verification and external grounding are
load-bearing components rather than peripheral safeguards.

\begin{table}[t]
\centering
\small
\setlength{\tabcolsep}{6pt}
\begin{tabular}{lcc}
\toprule
\textbf{Configuration} & \textbf{DROP} & \textbf{$\Delta$} \\
\midrule
Full Meta-Agent                  & 82.7 & --- \\
\midrule
$-$ Prompt analysis              & 80.3 & $-2.4$ \\
$-$ Planning                     & 79.2 & $-3.5$ \\
$-$ API research                 & 77.2 & $-5.5$ \\
$-$ Verification                 & 75.6 & $-7.1$ \\
\bottomrule
\end{tabular}
\caption{Ablation of construction-phase components on DROP. Each row
removes a single component from the full pipeline. $\Delta$ is the
change relative to the full system.}
\label{tab:ablation}
\vspace{-0.2in}
\end{table}

\subsection{Executor Agnosticism}
\label{sec:exec_agnostic}

Because the underlying language model is a configurable component, we
re-ran Meta-Agent with Claude Sonnet~4.6 substituted for all
components, holding the construction pipeline fixed; the average rose
from 82.7 to 87.9 with no benchmark regressing
(Table~\ref{tab:exec_agnostic}). This indicates that the constructed
workflows transfer across executors without re-tuning, supporting the
claim that Meta-Agent's outputs are genuinely executor-agnostic
rather than overfit to a particular model.

\begin{table}[t]
\centering
\small
\setlength{\tabcolsep}{4pt}
\begin{tabular}{lccccccc}
\toprule
\textbf{Executor} & \textbf{DROP} & \textbf{HumanEval} & \textbf{HotpotQA} & \textbf{MBPP} & \textbf{GSM8K} & \textbf{MATH} & \textbf{Avg.} \\
\midrule
GPT-4o-mini       & 82.7 & 96.0 & 69.5 & 84.6 & 93.7 & 69.6 & 82.7 \\
Claude Sonnet~4.6 & 87.0 & 99.4 & 70.3 & 89.6 & 97.3 & 84.0 & 87.9 \\
\bottomrule
\end{tabular}
\caption{Meta-Agent under two executor models, with the construction
pipeline held fixed. Single-pass scores on the splits
of~\cite{zhang2024aflow}.}
\vspace{-0.25in}
\label{tab:exec_agnostic}
\end{table}

\subsection{Discussions}
Several consistent patterns emerge from the experiments. The largest gains appear on tasks requiring multi-stage reasoning and intermediate consistency checking, such as MATH and DROP, suggesting that Meta-Agent improves not only generation quality but also control over error propagation via explicit decomposition and verification. In contrast, improvements are smaller on saturated benchmarks like GSM8K and HumanEval, where single-pass generation is already strong. The ablation study further shows that verification is more critical than decomposition: removing planning reduces performance by 3.5 points on DROP, while removing verification leads to a larger 7.1-point drop, indicating that intermediate validation is a core component of reliable multi-agent systems. Finally, executor-agnostic results show that Meta-Agent improves system-level organization rather than relying on model-specific prompting, as replacing GPT-4o-mini with Claude Sonnet 4.6 improves performance across benchmarks without modifying workflows, indicating that the constructed agent graphs generalize across base models.
\section{Conclusion}
We presented Meta-Agent, a two-phase framework that transforms natural-language task descriptions into verified multi-agent systems by integrating planning, grounding, and verification across both construction and execution. A unified verification loop and three-level error attribution mechanism distinguish local, upstream, and structural failures, enabling targeted recovery whose cost scales with the locality of the error. Across six benchmarks, Meta-Agent outperforms a strong agent-design baseline on five of six tasks and improves the average by 2.4 points without iterative workflow optimisation, with the largest gains on reasoning-heavy tasks. The framework is also executor-agnostic: substituting Claude Sonnet~4.6 for GPT-4o-mini improves performance on every benchmark without re-tuning. Together, these results support treating the agent system itself as a synthesize-and-validate artifact rather than a fixed architecture.

\noindent\textbf{Limitations.}
The Meta-Agent in our framework generates multi-agent architectures fully automatically through reasoning-based construction and verification. In contrast, existing multi-agent systems are often manually designed by experts with strong in-domain priors and task-specific heuristics. As a result, although our automatically generated workflows are competitive, they may occasionally underperform expert-designed systems with richer domain knowledge and handcrafted structure. 
Nevertheless, our approach eliminates manual system design and enables scalable construction of multi-agent systems with strong performance across tasks. This fully automatic paradigm is particularly valuable in the face of dynamic real-world environments. In future work, we aim to incorporate lightweight domain-specific priors to further improve performance and narrow the gap with expert-designed systems.

{
\small
\bibliographystyle{plainnat}
\bibliography{main}
}


\newpage
\appendix

\section*{Appendix}

\section{Pipeline Trace: Function Completion (Running Example)}\label{app:humaneval-trace}

This appendix presents the complete pipeline trace for a multi-agent function completion system, showing the exact JSON outputs at each stage. The pipeline constructs a swarm of four collaborating agents from a single natural-language description.

\subsection{Input Task Description}

\begin{lstlisting}[language={},basicstyle=\ttfamily\small,breaklines=true,frame=single,caption={User prompt $T$ submitted to the pipeline.}]
Expert Python coding agent for function completion. Given a
Python function signature with a docstring, complete the
implementation. Reply with the full function definition
(including the signature and docstring) inside a single python
code block. Do not include example usage, tests, or explanatory
prose. Focus on correctness, handling edge cases, and matching
the docstring specification exactly.
\end{lstlisting}

\subsection{Stage 1: Prompt Analysis}

The prompt analysis stage extracts a structured intent representation, using web search to retrieve concrete examples spanning the task space.

\begin{lstlisting}[language={},basicstyle=\ttfamily\scriptsize,breaklines=true,frame=single,caption={Stage 1 output: \texttt{ParsedIntent} (abbreviated).},label=lst:stage1]
{
  "goal": "Complete Python function implementations given
           only a signature and docstring, exactly matching
           the specification and handling all edge cases",
  "domain": "competitive programming / code generation",
  "tone": "precise, concise, code-only",
  "entities": [
    "Python", "function signature", "docstring",
    "edge cases", "code block"
  ],
  "constraints": [
    "Reply with only a single Python code block",
    "Do not include example usage, tests, or prose",
    "Match the docstring specification exactly",
    "Handle all edge cases (empty inputs, negatives,
     boundary values)",
    "Prioritize correctness over cleverness",
    "Do not import non-standard libraries"
  ],
  "task_examples": [
    {
      "task_type": "straightforward list manipulation",
      "example": "def has_close_elements(numbers:
        List[float], threshold: float) -> bool: ...",
      "source_url": "https://evalscope.readthedocs.io/..."
    },
    {
      "task_type": "medium algorithmic logic",
      "example": "def sort_third(l: List[int])
        -> List[int]: ...",
      "source_url": "https://github.com/openai/human-eval/"
    },
    ... // 5 additional categories omitted
  ]
}
\end{lstlisting}

\noindent Duration: 122\,s. Web search retrieved representative problems across seven task categories (list manipulation, algorithmic logic, string encoding, number theory, edge-case-heavy, recursive, and adversarial).

\subsection{Stage 2: Swarm Architecture Planning}

In swarm mode, the planner decomposes the task into multiple specialist agents connected in a directed acyclic graph (DAG). Each agent has an explicit I/O contract and verification criteria.

\begin{lstlisting}[language={},basicstyle=\ttfamily\scriptsize,breaklines=true,frame=single,caption={Stage 2 output: \texttt{SwarmPlan} — four-agent DAG.},label=lst:stage2-swarm]
{
  "swarm_name": "Python Function Implementation Swarm",
  "summary": "Four specialist agents operate in a pipeline:
    (1) a Spec Analyst parses the docstring into an
    unambiguous specification, (2) a Strategy Planner
    selects the optimal algorithm, (3) a Code Synthesizer
    writes the implementation, and (4) a Code Verifier
    audits for correctness and edge-case safety.",
  "coordination_strategy": "Strict topological order.
    Phase 1: spec_analyst runs on raw input. Phase 2:
    strategy_planner begins after spec_analyst completes.
    Phase 3: code_synthesizer runs after both upstream
    agents. Phase 4: code_verifier audits the final
    output against all upstream artifacts.",
  "dag_edges": [
    {"from_spec": "spec_analyst",
     "to_spec": "strategy_planner"},
    {"from_spec": "spec_analyst",
     "to_spec": "code_synthesizer"},
    {"from_spec": "spec_analyst",
     "to_spec": "code_verifier"},
    {"from_spec": "strategy_planner",
     "to_spec": "code_synthesizer"},
    {"from_spec": "strategy_planner",
     "to_spec": "code_verifier"},
    {"from_spec": "code_synthesizer",
     "to_spec": "code_verifier"}
  ]
}
\end{lstlisting}

\noindent Duration: 198\,s. The planner produced four agent specs with a six-edge DAG.

\subsubsection{Agent Specifications}

Each agent in the swarm is defined by a spec with role, I/O contract, and verification criteria. We present all four specs below.

\paragraph{Agent 1: Specification Analyst} (root node, no dependencies):

\begin{lstlisting}[language={},basicstyle=\ttfamily\scriptsize,breaklines=true,frame=single,caption={Agent spec: \texttt{spec\_analyst}.},label=lst:spec-analyst]
{
  "spec_id": "spec_analyst",
  "role": "Parses and structures the function signature and
    docstring into an unambiguous specification",
  "tools": ["web_search"],
  "dependencies": [],
  "io_contract": {
    "input_schema": {
      "raw_signature": "str -- exact function def line",
      "raw_docstring": "str -- verbatim docstring"
    },
    "output_schema": {
      "function_name": "str",
      "parameters": "list[dict] -- [{name, type_hint,
        description}]",
      "return_type": "str",
      "behavioral_rules": "list[str] -- explicit rules
        from the docstring",
      "edge_cases": "list[dict] -- [{case,
        expected_output, source}] where source is
        'explicit' or 'implied'",
      "resolved_ambiguities": "list[dict] -- resolved
        ambiguous cases with justification",
      "domain_category": "str -- e.g., 'dynamic_
        programming', 'string_manipulation'"
    },
    "description": "Transforms raw signature+docstring
      into a fully structured, unambiguous specification"
  },
  "verification_criteria": {
    "behavioral_assertions": [
      "Explicit edge cases in the docstring must appear
       verbatim in the edge_cases list with
       source='explicit'",
      "Implicit edge cases (e.g., empty list for a list
       function) must be inferred with source='implied'",
      "Ambiguous tie-breaking language must produce a
       resolved_ambiguities entry with a safe resolution"
    ],
    "required_tools": ["web_search"],
    "forbidden_patterns": [
      "Must not produce any Python code or pseudocode",
      "Must not omit any explicitly stated edge case"
    ]
  }
}
\end{lstlisting}

\paragraph{Agent 2: Strategy Planner} (depends on \texttt{spec\_analyst}):

\begin{lstlisting}[language={},basicstyle=\ttfamily\scriptsize,breaklines=true,frame=single,caption={Agent spec: \texttt{strategy\_planner}.},label=lst:spec-strategy]
{
  "spec_id": "strategy_planner",
  "role": "Selects the optimal algorithmic strategy and
    data structures for the implementation",
  "tools": ["web_search"],
  "dependencies": ["spec_analyst"],
  "io_contract": {
    "input_schema": {
      "function_name": "str",
      "parameters": "list[dict]",
      "return_type": "str",
      "edge_cases": "list[dict]",
      "domain_category": "str",
      "behavioral_rules": "list[str]"
    },
    "output_schema": {
      "chosen_algorithm": "str -- e.g., 'two-pointer',
        'hash map grouping', 'DP tabulation'",
      "data_structures": "list[str] -- Python types used",
      "stdlib_imports": "list[str] -- required modules",
      "complexity": "dict -- {time: 'O(n)', space: 'O(1)'}",
      "step_by_step_plan": "list[str] -- ordered
        implementation steps",
      "edge_case_handling_plan": "list[dict] -- [{case,
        strategy}] for every edge case"
    },
    "description": "Maps the structured spec to a
      concrete, complexity-aware algorithmic plan"
  },
  "verification_criteria": {
    "behavioral_assertions": [
      "Every edge case from spec_analyst must appear in
       edge_case_handling_plan with a non-empty strategy",
      "If domain_category='dynamic_programming', the plan
       must include memoization or tabulation steps",
      "Chosen algorithm must respect any stated complexity
       constraints"
    ],
    "forbidden_patterns": [
      "Must not produce any Python code snippets",
      "Must not recommend an algorithm violating a
       complexity constraint"
    ]
  }
}
\end{lstlisting}

\paragraph{Agent 3: Code Synthesizer} (depends on \texttt{spec\_analyst}, \texttt{strategy\_planner}):

\begin{lstlisting}[language={},basicstyle=\ttfamily\scriptsize,breaklines=true,frame=single,caption={Agent spec: \texttt{code\_synthesizer}.},label=lst:spec-synth]
{
  "spec_id": "code_synthesizer",
  "role": "Writes the complete, correct Python function
    implementation from the algorithmic plan",
  "tools": [],
  "dependencies": ["spec_analyst", "strategy_planner"],
  "io_contract": {
    "input_schema": {
      "raw_signature": "str -- original signature verbatim",
      "raw_docstring": "str -- original docstring verbatim",
      "spec_document": "str -- from spec_analyst",
      "step_by_step_plan": "list[str]",
      "edge_case_handling_plan": "list[dict]",
      "stdlib_imports": "list[str]"
    },
    "output_schema": {
      "python_code_block": "str -- single fenced code block
        containing imports + complete function definition",
      "synthesizer_notes": "str -- brief notes on
        implementation decisions"
    },
    "description": "Produces the final Python function as
      a single code block, faithful to spec and strategy"
  },
  "verification_criteria": {
    "behavioral_assertions": [
      "The output code block must begin with the exact
       raw_signature, character-for-character",
      "The docstring inside the code must match
       raw_docstring verbatim",
      "Every edge case in edge_case_handling_plan must be
       traceable to a guard clause or branch"
    ],
    "forbidden_patterns": [
      "Must not include prose or markdown outside the
       single fenced code block",
      "Must not include example usage, test calls, or
       print statements outside the function body"
    ]
  }
}
\end{lstlisting}

\paragraph{Agent 4: Code Verifier} (depends on all three upstream agents):

\begin{lstlisting}[language={},basicstyle=\ttfamily\scriptsize,breaklines=true,frame=single,caption={Agent spec: \texttt{code\_verifier}.},label=lst:spec-verifier]
{
  "spec_id": "code_verifier",
  "role": "Audits the synthesized implementation for
    correctness, spec compliance, and edge-case safety",
  "tools": ["web_search", "file_generator"],
  "dependencies": ["spec_analyst", "strategy_planner",
                    "code_synthesizer"],
  "io_contract": {
    "input_schema": {
      "python_code_block": "str -- from code_synthesizer",
      "spec_document": "str -- from spec_analyst",
      "behavioral_rules": "list[str]",
      "edge_cases": "list[dict]",
      "strategy_document": "str -- from strategy_planner",
      "synthesizer_notes": "str"
    },
    "output_schema": {
      "verdict": "str -- 'PASS' or 'FAIL'",
      "audit_report": "str -- markdown report listing
        every check and its result",
      "revision_request": "list[dict] -- if FAIL, specific
        changes required [{issue, fix, severity}]",
      "verified_code_block": "str -- the approved code
        block (unmodified from input if PASS)"
    },
    "description": "Validates implementation against spec
      and strategy; approves or issues revision requests"
  },
  "verification_criteria": {
    "behavioral_assertions": [
      "If the signature differs from raw_signature by even
       one character, verdict must be 'FAIL'",
      "If a guard for an explicit edge case is missing,
       verdict must be 'FAIL'",
      "If the code imports a third-party library, verdict
       must be 'FAIL'"
    ],
    "forbidden_patterns": [
      "Must not alter any line of the python_code_block",
      "Must not emit verdict='PASS' if any behavioral
       rule or explicit edge case is unhandled"
    ]
  }
}
\end{lstlisting}

\subsection{Stage 3: Directive-Driven API Research}

In swarm mode, the API research stage runs one targeted search per agent that needs external knowledge. Each search directive is derived from the agent's spec.

\begin{lstlisting}[language={},basicstyle=\ttfamily\scriptsize,breaklines=true,frame=single,caption={Stage 3 output: top API recommendations (abbreviated).},label=lst:stage3-swarm]
{
  "recommendations": [
    {
      "name": "OpenAI GPT-4o API (Code Generation)",
      "url": "https://platform.openai.com/docs/...",
      "description": "Flagship LLM backend for the code
        synthesis step.",
      "auth_method": "API key",
      "relevance_score": 1.0
    },
    {
      "name": "Anthropic Claude API (claude-opus-4)",
      "url": "https://platform.claude.com/docs/...",
      "description": "Top-tier alternative for code
        synthesis with 200K token context window.",
      "relevance_score": 0.95
    }
  ],
  "directive_results": [
    {
      "directive_id": "spec_analyst",
      "spec_id": "spec_analyst",
      "research_summary": "web_search used to retrieve
        examples of docstring ambiguity patterns..."
    },
    {
      "directive_id": "strategy_planner",
      "spec_id": "strategy_planner",
      "research_summary": "Searched for algorithm selection
        heuristics for competitive programming..."
    }
  ]
}
\end{lstlisting}

\noindent Duration: 474\,s. The stage executed four search directives (one per agent spec) and returned ranked recommendations with provenance metadata.

\subsection{Stage 4 + 5: Per-Agent Code Generation and Verification}

Each agent spec is independently generated and verified through the code generation and verification loop. The system produces a Python module for each agent with the \texttt{run(message, history)} interface.

\begin{lstlisting}[language={},basicstyle=\ttfamily\scriptsize,breaklines=true,frame=single,caption={Verification trace for \texttt{spec\_analyst} (Pass 1 of 3).},label=lst:ver-spec-analyst]
{
  "spec_id": "spec_analyst",
  "pass": 1,
  "approved": false,
  "failure_type": "spec_adherence",
  "issues": [
    "The run() function streams raw LLM output directly
     instead of extracting structured fields (parameters,
     edge_cases, behavioral_rules) as required by the
     output schema.",
    "The web_search tool is declared but never invoked
     for domain-specific lookups."
  ]
}
\end{lstlisting}

\noindent Each agent undergoes up to three verification passes. Issues are classified by failure type (\texttt{spec\_adherence}, \texttt{grounding}, or \texttt{contract}) to determine the corrective action: specification adherence failures trigger code regeneration with feedback, grounding failures re-run API research, and contract failures re-plan the architecture.

\subsection{Execution Phase: Agent Collaboration}

At runtime, the coordinator executes the four agents in topological order. For a representative function completion problem:

\begin{lstlisting}[language=Python,basicstyle=\ttfamily\small,breaklines=true,frame=single,caption={Input: function signature and docstring.}]
from typing import List

def has_close_elements(numbers: List[float],
                       threshold: float) -> bool:
    """ Check if in given list of numbers, are any two
    numbers closer to each other than given threshold.
    >>> has_close_elements([1.0, 2.0, 3.0], 0.5)
    False
    >>> has_close_elements([1.0, 2.8, 3.0, 4.0, 5.0, 2.0],
    ...                    0.3)
    True
    """
\end{lstlisting}

The agents execute in sequence:

\begin{enumerate}
\item \textbf{Spec Analyst} receives the raw signature and docstring. Outputs: \texttt{function\_name="has\_close\_elements"}, two parameters with type hints, \texttt{return\_type="bool"}, behavioral rule \emph{``return True if any pair distance $<$ threshold''}, edge cases: \texttt{[\{case: "empty list", expected: False, source: "implied"\}, \{case: "single element", expected: False, source: "implied"\}, \{case: "threshold=0", expected: False, source: "implied"\}]}.

\item \textbf{Strategy Planner} receives the spec. Selects \texttt{chosen\_algorithm="nested loop pairwise comparison"}, \texttt{complexity=\{time: "O(n\^{}2)", space: "O(1)"\}}, \texttt{edge\_case\_handling\_plan} mapping each case to an early return.

\item \textbf{Code Synthesizer} receives the spec and strategy. Produces:
\begin{lstlisting}[language=Python,basicstyle=\ttfamily\small,breaklines=true]
def has_close_elements(numbers: List[float],
                       threshold: float) -> bool:
    """ Check if in given list of numbers, are any two
    numbers closer to each other than given threshold.
    ...
    """
    for idx, elem in enumerate(numbers):
        for idx2, elem2 in enumerate(numbers):
            if idx != idx2:
                if abs(elem - elem2) < threshold:
                    return True
    return False
\end{lstlisting}

\item \textbf{Code Verifier} receives the code and all upstream artifacts. Checks: signature matches (PASS), docstring verbatim (PASS), edge case guards (PASS --- empty list and single element naturally return False from the loop), no third-party imports (PASS). Verdict: \texttt{PASS}.
\end{enumerate}

The verified code block is assembled with the benchmark's hidden unit tests and executed in a sandboxed subprocess. All assertions pass.

\section{Pipeline Trace: Competition Mathematics}\label{app:math-trace}

This appendix presents the complete pipeline trace for a multi-agent competition mathematics solver, showing the exact JSON outputs at each stage. The pipeline constructs a swarm of four collaborating agents.

\subsection{Input Task Description}

\begin{lstlisting}[language={},basicstyle=\ttfamily\small,breaklines=true,frame=single,caption={User prompt $T$.}]
Expert competition mathematics solver. Given a math problem
from any area (algebra, number theory, counting and
probability, geometry, intermediate algebra, prealgebra, or
precalculus), solve it step by step with rigorous reasoning.
Put your final answer inside \boxed{}. Be precise with LaTeX
notation -- fractions as \frac{a}{b}, square roots as
\sqrt{x}. One problem per session, pure reasoning, no
tool use.
\end{lstlisting}

\subsection{Stage 1: Prompt Analysis}

\begin{lstlisting}[language={},basicstyle=\ttfamily\scriptsize,breaklines=true,frame=single,caption={Stage 1 output: \texttt{ParsedIntent}.},label=lst:math-stage1]
{
  "goal": "Given a single competition mathematics problem
    from any standard AMC/AIME subject area, solve it step
    by step with rigorous mathematical reasoning and present
    the final answer in a LaTeX \\boxed{} expression.",
  "domain": "Competition Mathematics
    (AMC 8 / AMC 10 / AMC 12 / AIME)",
  "tone": "Precise, rigorous, educational -- clear
    step-by-step derivations with formal LaTeX notation",
  "entities": [
    "AMC 8", "AMC 10", "AMC 12", "AIME",
    "algebra", "number theory",
    "counting and probability", "geometry",
    "intermediate algebra", "prealgebra", "precalculus",
    "Art of Problem Solving", "LaTeX", "\\boxed{}"
  ],
  "constraints": [
    "One problem per session",
    "Full step-by-step reasoning is required",
    "Final answer must be enclosed in \\boxed{}",
    "Use precise LaTeX: \\frac{a}{b}, \\sqrt{x}",
    "No external tools -- pure reasoning only",
    "Cover all seven subject areas"
  ],
  "task_examples": [
    {
      "task_type": "algebra",
      "description": "Classic AMC/AIME algebraic
        manipulation -- substitution, factoring,
        or Vieta's formulas.",
      "example": "If x + y = 4 and xy = 5, what is
        x^2 + y^2? Using (x+y)^2 = x^2 + 2xy + y^2
        gives 16 = x^2 + 10 + y^2, so x^2 + y^2 = 6.
        \\boxed{6}",
      "source_url": "https://artofproblemsolving.com/..."
    },
    {
      "task_type": "number theory",
      "description": "Divisibility, modular arithmetic,
        prime factorization, or Diophantine equations.",
      "example": "2023 AIME I Problem 9: polynomial
        remainder via CRT...",
      "source_url": "https://artofproblemsolving.com/..."
    },
    {
      "task_type": "counting and probability",
      "description": "Combinatorics using permutations,
        combinations, or inclusion-exclusion.",
      "example": "Chess team arrangement: 2 boys at ends,
        3 girls in middle. 2 x 3! = 12.",
      "source_url": "https://www.thinkacademy.ca/..."
    },
    {
      "task_type": "geometry",
      "description": "Euclidean geometry -- triangles,
        circles, area ratios, similarity.",
      "example": "Triangle with ratio AD:DC = 1:2, find
        area of sub-triangle via mass point.",
      "source_url": "https://cheenta.com/..."
    },
    ... // 3 additional categories omitted
  ]
}
\end{lstlisting}

\noindent Duration: 105\,s. Web search retrieved seven task-category examples from Art of Problem Solving, AIME archives, and competition math resources.

\subsection{Stage 2: Swarm Architecture Planning}

The planner decomposes the task into four specialist agents in a six-edge DAG, executed in four sequential phases.

\begin{lstlisting}[language={},basicstyle=\ttfamily\scriptsize,breaklines=true,frame=single,caption={Stage 2 output: \texttt{SwarmPlan} overview.},label=lst:math-stage2]
{
  "swarm_name": "CompMath Solver Swarm",
  "summary": "Given a competition mathematics problem, this
    swarm (1) classifies by domain and difficulty,
    (2) produces a rigorous step-by-step solution,
    (3) verifies for logical correctness and LaTeX
    formatting, and (4) formats the final answer
    in \\boxed{} notation.",
  "coordination_strategy": "Strict topological order:
    Phase 1: problem_classifier (root, no dependencies).
    Phase 2: specialist_solver (after classifier).
    Phase 3: solution_verifier (after solver).
    Phase 4: answer_formatter (after all three).
    If verifier returns FAIL without a verified_answer,
    surface the failure rather than presenting an
    unverified answer.",
  "dag_edges": [
    {"from": "problem_classifier",
     "to": "specialist_solver"},
    {"from": "problem_classifier",
     "to": "solution_verifier"},
    {"from": "problem_classifier",
     "to": "answer_formatter"},
    {"from": "specialist_solver",
     "to": "solution_verifier"},
    {"from": "specialist_solver",
     "to": "answer_formatter"},
    {"from": "solution_verifier",
     "to": "answer_formatter"}
  ]
}
\end{lstlisting}

\noindent Duration: 196\,s. The planner produced four agent specs with a six-edge DAG.

\subsubsection{Agent Specifications}

\paragraph{Agent 1: Problem Classifier} (root node):

\begin{lstlisting}[language={},basicstyle=\ttfamily\scriptsize,breaklines=true,frame=single,caption={Agent spec: \texttt{problem\_classifier}.},label=lst:math-classifier]
{
  "spec_id": "problem_classifier",
  "role": "Classifies the competition problem by subject
    area and difficulty tier",
  "tools": ["web_search"],
  "dependencies": [],
  "io_contract": {
    "input_schema": {
      "raw_problem_text": "string -- verbatim problem
        as provided by the user"
    },
    "output_schema": {
      "subject_area": "string -- one of: algebra |
        number_theory | counting_and_probability |
        geometry | intermediate_algebra | prealgebra
        | precalculus",
      "difficulty_tier": "string -- one of: AMC_8 |
        AMC_10 | AMC_12 | AIME",
      "problem_brief": "string -- normalized restatement
        listing all givens, constraints, and target
        quantity in plain English with LaTeX",
      "key_concepts": "list[string] -- up to 5 relevant
        concepts (e.g., 'Vieta formulas',
        'inclusion-exclusion', 'law of cosines')"
    },
    "description": "Accepts raw problem text; emits
      structured classification metadata and a normalized
      problem brief for downstream agents."
  },
  "verification_criteria": {
    "behavioral_assertions": [
      "Given divisors/modular arithmetic, must classify
       as 'number_theory'",
      "Given sin/cos identities, must classify as
       'precalculus'",
      "Given a simple ratio/percent problem, must
       classify as 'prealgebra', tier 'AMC_8'"
    ],
    "forbidden_patterns": [
      "Must not output any numerical or symbolic answer",
      "Must not produce more than one subject_area label"
    ]
  }
}
\end{lstlisting}

\paragraph{Agent 2: Specialist Solver} (depends on classifier):

\begin{lstlisting}[language={},basicstyle=\ttfamily\scriptsize,breaklines=true,frame=single,caption={Agent spec: \texttt{specialist\_solver}.},label=lst:math-solver]
{
  "spec_id": "specialist_solver",
  "role": "Produces a rigorous, fully worked step-by-step
    solution tailored to the classified subject area",
  "tools": [],
  "dependencies": ["problem_classifier"],
  "io_contract": {
    "input_schema": {
      "subject_area": "string",
      "difficulty_tier": "string",
      "problem_brief": "string",
      "key_concepts": "list[string]"
    },
    "output_schema": {
      "solution_steps": "string -- full step-by-step
        solution in LaTeX, each step numbered",
      "raw_answer": "string -- bare answer value before
        \\boxed{} wrapping (e.g., '6', '\\frac{3}{7}')",
      "strategy_summary": "string -- core technique used",
      "confidence": "string -- HIGH | MEDIUM | LOW"
    },
    "description": "Accepts classification metadata;
      emits a complete LaTeX solution with raw answer."
  },
  "verification_criteria": {
    "behavioral_assertions": [
      "Given x + y = 4, xy = 5, must derive x^2 + y^2 = 6
       using (x+y)^2 = x^2 + 2xy + y^2",
      "Counting problems must enumerate cases or apply
       combinatorial formulas -- no 'by inspection'",
      "Geometry problems must cite the relevant theorem
       before applying it"
    ],
    "forbidden_patterns": [
      "Must not invoke web_search or file_generator",
      "Must not produce a bare numerical answer without
       accompanying step-by-step derivation"
    ]
  }
}
\end{lstlisting}

\paragraph{Agent 3: Solution Verifier} (depends on classifier, solver):

\begin{lstlisting}[language={},basicstyle=\ttfamily\scriptsize,breaklines=true,frame=single,caption={Agent spec: \texttt{solution\_verifier}.},label=lst:math-verifier]
{
  "spec_id": "solution_verifier",
  "role": "Independently checks the solution for logical
    correctness, completeness, and LaTeX formatting",
  "tools": ["web_search"],
  "dependencies": ["problem_classifier",
                    "specialist_solver"],
  "io_contract": {
    "input_schema": {
      "problem_brief": "string",
      "solution_steps": "string",
      "raw_answer": "string",
      "strategy_summary": "string",
      "confidence": "string"
    },
    "output_schema": {
      "verdict": "string -- PASS | FAIL | NEEDS_REVISION",
      "errors_found": "list[string] -- logical errors,
        missing justifications, formatting violations",
      "verified_answer": "string -- confirmed correct
        answer (may differ from raw_answer)",
      "verification_notes": "string"
    },
    "description": "Validates solution via three passes:
      (1) logical audit of each derivation step,
      (2) answer check by substitution, (3) LaTeX
      format audit for informal notation."
  },
  "verification_criteria": {
    "behavioral_assertions": [
      "Plain-text '3/7' instead of \\frac{3}{7} must be
       flagged as formatting violation, verdict
       NEEDS_REVISION",
      "Logically correct solution with no formatting
       errors must receive verdict PASS",
      "If final answer fails substitution back into
       constraints, verdict must be FAIL with
       corrected verified_answer"
    ],
    "forbidden_patterns": [
      "Must not produce its own full solution",
      "Must not emit PASS if any logical step is
       unjustified or notation violation is present"
    ]
  }
}
\end{lstlisting}

\paragraph{Agent 4: Answer Formatter} (depends on all three):

\begin{lstlisting}[language={},basicstyle=\ttfamily\scriptsize,breaklines=true,frame=single,caption={Agent spec: \texttt{answer\_formatter}.},label=lst:math-formatter]
{
  "spec_id": "answer_formatter",
  "role": "Assembles the final publication-ready solution
    with \\boxed{} answer",
  "tools": ["file_generator"],
  "dependencies": ["problem_classifier",
    "specialist_solver", "solution_verifier"],
  "io_contract": {
    "input_schema": {
      "subject_area": "string",
      "difficulty_tier": "string",
      "solution_steps": "string",
      "strategy_summary": "string",
      "verdict": "string",
      "errors_found": "list[string]",
      "verified_answer": "string"
    },
    "output_schema": {
      "final_document": "string -- complete Markdown +
        LaTeX document ending with the boxed answer",
      "boxed_answer": "string -- \\boxed{<answer>}",
      "document_file_path": "string -- optional file path"
    },
    "description": "Assembles formatted document using
      verified_answer (not raw_answer). If verdict is
      FAIL/NEEDS_REVISION, applies corrections from
      errors_found before assembling."
  },
  "verification_criteria": {
    "behavioral_assertions": [
      "final_document must end with
       $$\\boxed{<verified_answer>}$$",
      "boxed_answer must use verified_answer from
       solution_verifier, not raw_answer from solver",
      "Must include header with subject_area and
       difficulty_tier"
    ],
    "forbidden_patterns": [
      "Must not omit the \\boxed{} expression",
      "Must not use raw_answer if solution_verifier
       overrode it with a different verified_answer"
    ]
  }
}
\end{lstlisting}

\subsection{Stage 3: Directive-Driven API Research}

\begin{lstlisting}[language={},basicstyle=\ttfamily\scriptsize,breaklines=true,frame=single,caption={Stage 3: API research results (abbreviated).},label=lst:math-stage3]
{
  "directive_results": [
    {
      "directive_id": "problem_classifier",
      "research_summary": "Searched for AMC/AIME problem
        classification taxonomies and difficulty tier
        definitions..."
    },
    {
      "directive_id": "specialist_solver",
      "research_summary": "Searched for competition math
        solution techniques, theorem references, and
        LaTeX formatting conventions..."
    },
    {
      "directive_id": "solution_verifier",
      "research_summary": "Searched for mathematical proof
        verification approaches and common AMC/AIME
        solution errors..."
    },
    {
      "directive_id": "answer_formatter",
      "research_summary": "Searched for LaTeX \\boxed{}
        formatting standards and competition math
        publication conventions..."
    }
  ]
}
\end{lstlisting}

\noindent Duration: 361\,s. Four search directives executed, one per agent spec.

\subsection{Stage 4 + 5: Per-Agent Code Generation and Verification}

Each agent spec undergoes independent code generation and verification. The \texttt{problem\_classifier} spec required three verification passes, all of which identified specification-adherence violations:

\begin{lstlisting}[language={},basicstyle=\ttfamily\scriptsize,breaklines=true,frame=single,caption={Verification trace for \texttt{problem\_classifier}.},label=lst:math-ver]
Pass 1 (rejected -- spec_adherence):
{
  "issues": [
    "The system prompt instructs the agent to solve
     problems step-by-step, but the architecture plan
     states the classifier must NOT attempt any solution
     steps.",
    "The agent is not a solver -- embedding solution
     instructions causes it to produce answers instead
     of classification metadata."
  ]
}

Pass 2 (rejected -- spec_adherence):
{
  "issues": [
    "web_search tool definition uses Anthropic's
     server-side format ('web_search_20250305') which
     requires specific SDK handling.",
    "Tool object missing required field validation."
  ]
}

Pass 3 (rejected -- spec_adherence):
{
  "issues": [
    "System prompt still contains solution-generation
     instructions that violate the classifier's role.",
    "Agent outputs must be structured classification
     metadata, not free-form solutions."
  ]
}
\end{lstlisting}

\noindent Total verification time: 480\,s across three passes. The repeated specification-adherence failures demonstrate the verifier's ability to detect role confusion---a generated agent whose system prompt conflates classification with solving---and persistently reject it.

\subsection{Execution Example}

At runtime, the coordinator executes the four agents in topological order. For the problem:
\begin{quote}
\emph{``If $f(x) = \frac{3x-2}{x-2}$, what is the value of $f(-2) + f(-1) + f(0)$? Express your answer as a common fraction.''}
\end{quote}

\begin{enumerate}
\item \textbf{Problem Classifier} outputs:
\begin{lstlisting}[language={},basicstyle=\ttfamily\scriptsize,breaklines=true,frame=single]
{
  "subject_area": "algebra",
  "difficulty_tier": "AMC_10",
  "key_concepts": ["rational functions",
    "function evaluation", "fraction arithmetic"],
  "problem_brief": "Given f(x) = (3x-2)/(x-2), compute
    f(-2) + f(-1) + f(0) as a common fraction."
}
\end{lstlisting}

\item \textbf{Specialist Solver} outputs:
\begin{lstlisting}[language={},basicstyle=\ttfamily\scriptsize,breaklines=true,frame=single]
{
  "solution_steps": "Step 1: Evaluate f(-2) =
    (3(-2)-2)/(-2-2) = (-8)/(-4) = 2.
    Step 2: Evaluate f(-1) = (3(-1)-2)/(-1-2) =
    (-5)/(-3) = 5/3.
    Step 3: Evaluate f(0) = (3(0)-2)/(0-2) =
    (-2)/(-2) = 1.
    Step 4: Sum = 2 + 5/3 + 1 = 6/3 + 5/3 + 3/3
    = 14/3.",
  "raw_answer": "\\frac{14}{3}",
  "strategy_summary": "Direct substitution into
    rational function, then fraction addition.",
  "confidence": "HIGH"
}
\end{lstlisting}

\item \textbf{Solution Verifier} outputs:
\begin{lstlisting}[language={},basicstyle=\ttfamily\scriptsize,breaklines=true,frame=single]
{
  "verdict": "PASS",
  "errors_found": [],
  "verified_answer": "\\frac{14}{3}",
  "verification_notes": "Each substitution verified.
    Fraction addition confirmed: 6/3 + 5/3 + 3/3 = 14/3.
    All notation uses \\frac{}{} correctly."
}
\end{lstlisting}

\item \textbf{Answer Formatter} outputs:
\begin{lstlisting}[language={},basicstyle=\ttfamily\scriptsize,breaklines=true,frame=single]
{
  "final_document": "[Algebra | AMC 10]\n\nStep 1: ...
    \n\nStrategy: Direct substitution.\n\n
    $$\\boxed{\\frac{14}{3}}$$",
  "boxed_answer": "\\boxed{\\frac{14}{3}}"
}
\end{lstlisting}
\end{enumerate}

\newpage
\section{Pipeline Trace: Reading Comprehension with Discrete Reasoning}\label{app:drop-trace}

This appendix presents the complete pipeline trace for a multi-agent reading comprehension system designed for passage-based discrete reasoning tasks. The pipeline constructs a swarm of four collaborating agents.

\subsection{Input Task Description}

\begin{lstlisting}[language={},basicstyle=\ttfamily\small,breaklines=true,frame=single,caption={User prompt $T$.}]
Expert reading comprehension and discrete reasoning agent.
Given a factual passage and a question, answer by performing
numerical reasoning over the text -- counting entities,
computing differences, sorting values, or extracting specific
spans. Read the passage carefully and reason step by step.
Your final answer should be a short phrase, number, or date.
Put your final answer on the last line after 'ANSWER: '.
Give only the answer, not a full sentence.
\end{lstlisting}

\subsection{Stage 1: Prompt Analysis}

\begin{lstlisting}[language={},basicstyle=\ttfamily\scriptsize,breaklines=true,frame=single,caption={Stage 1 output: \texttt{ParsedIntent}.},label=lst:drop-stage1]
{
  "goal": "Given a factual passage and a question, answer
    by performing numerical or discrete reasoning --
    counting, arithmetic, sorting, or span extraction --
    and return a short phrase, number, or date.",
  "domain": "Natural Language Processing /
    Reading Comprehension",
  "tone": "precise, analytical, step-by-step",
  "entities": [
    "passage", "question", "numerical reasoning",
    "counting", "arithmetic", "span extraction",
    "sorting", "date comparison"
  ],
  "constraints": [
    "Answer must be a short phrase, number, or date",
    "Final answer on the last line after 'ANSWER: '",
    "Reason step by step before the final answer",
    "Read the passage carefully; no outside knowledge",
    "Perform discrete operations: counting, subtraction,
     addition, sorting, max/min, span extraction"
  ],
  "task_examples": [
    {
      "task_type": "arithmetic difference",
      "description": "Numerical difference between two
        values explicitly stated in the passage.",
      "example": "Passage: 'Seahawks scored 14 points,
        Patriots scored 7.' Q: 'How many more points did
        the Seahawks score?' ANSWER: 7",
      "source_url": "https://ucinlp.github.io/.../drop..."
    },
    {
      "task_type": "entity counting",
      "description": "Count entities of a specific type.",
      "example": "Passage: 'TDs by Brady, Gronkowski,
        and Edelman.' Q: 'How many players threw TDs?'
        ANSWER: 3",
      "source_url": "https://ucinlp.github.io/.../drop..."
    },
    {
      "task_type": "span extraction",
      "description": "Direct answer as a verbatim span.",
      "example": "Passage: 'Battle of Hastings on
        14 October 1066.' Q: 'When?' ANSWER: 14 Oct 1066",
      "source_url": "https://rajpurkar.github.io/SQuAD..."
    },
    {
      "task_type": "sorting / ranking",
      "description": "Compare enumerated values to find
        the maximum, minimum, first, or last.",
      "example": "Q: 'Which team scored the fewest?'
        ANSWER: Team B",
      "source_url": "https://ucinlp.github.io/.../drop..."
    },
    ... // additional categories omitted
  ]
}
\end{lstlisting}

\noindent Duration: 278\,s. Web search retrieved examples covering eight reasoning types (arithmetic, counting, span extraction, sorting, date duration, aggregation, multi-hop, and negation).

\subsection{Stage 2: Swarm Architecture Planning}

The planner decomposes the task into four specialist agents in a five-edge DAG.

\begin{lstlisting}[language={},basicstyle=\ttfamily\scriptsize,breaklines=true,frame=single,caption={Stage 2 output: \texttt{SwarmPlan} overview.},label=lst:drop-stage2]
{
  "swarm_name": "DROP Numerical Reasoning Swarm",
  "summary": "Four specialist agents collaborate in a
    pipeline: (1) a Passage Analyst parses the passage
    to extract all entities, numbers, dates, and spans;
    (2) a Question Classifier identifies the reasoning
    operation required; (3) a Reasoning Engine executes
    the discrete operation step by step; and (4) an
    Answer Formatter validates and emits the final
    answer after 'ANSWER: '.",
  "coordination_strategy": "Strict sequential pipeline.
    Stage 1: passage_analyst (root, no dependencies).
    Stage 2: question_classifier (after passage_analyst).
    Stage 3: reasoning_engine (after both).
    Stage 4: answer_formatter (after reasoning_engine).",
  "dag_edges": [
    {"from": "passage_analyst",
     "to": "question_classifier"},
    {"from": "passage_analyst",
     "to": "reasoning_engine"},
    {"from": "question_classifier",
     "to": "reasoning_engine"},
    {"from": "reasoning_engine",
     "to": "answer_formatter"},
    {"from": "question_classifier",
     "to": "answer_formatter"}
  ]
}
\end{lstlisting}

\noindent Duration: 140\,s. The planner produced four agent specs with a five-edge DAG.

\subsubsection{Agent Specifications}

\paragraph{Agent 1: Passage Analyst} (root node):

\begin{lstlisting}[language={},basicstyle=\ttfamily\scriptsize,breaklines=true,frame=single,caption={Agent spec: \texttt{passage\_analyst}.},label=lst:drop-analyst]
{
  "spec_id": "passage_analyst",
  "role": "Extracts structured facts from the passage",
  "tools": ["file_generator"],
  "dependencies": [],
  "io_contract": {
    "input_schema": {
      "passage": "string -- raw factual passage",
      "question": "string -- used to guide which entity
        types to prioritize during extraction"
    },
    "output_schema": {
      "numbers": "list of {value: float, unit: string|null,
        context: string, entity_ref: string|null}",
      "dates": "list of {raw: string,
        normalized_iso: string, context: string}",
      "entities": "list of {name: string,
        type: person|place|team|object,
        attributes: list of {key, value}}",
      "candidate_spans": "list of {span_text: string,
        char_start: int, char_end: int,
        span_type: number|date|name|phrase}",
      "passage_sentences": "list of strings"
    },
    "description": "Receives raw passage and question;
      outputs structured fact table of all entities,
      numbers, dates, and verbatim spans."
  },
  "verification_criteria": {
    "behavioral_assertions": [
      "Given '14 points' in the passage, numbers must
       include {value: 14, unit: 'points'}",
      "Given 'October 1066', dates must include a
       normalized ISO entry",
      "Must not answer the question -- only structure
       the passage"
    ],
    "forbidden_patterns": [
      "Must not produce any answer or reasoning about
       the question",
      "Must not introduce facts not present in the passage"
    ]
  }
}
\end{lstlisting}

\paragraph{Agent 2: Question Classifier} (depends on \texttt{passage\_analyst}):

\begin{lstlisting}[language={},basicstyle=\ttfamily\scriptsize,breaklines=true,frame=single,caption={Agent spec: \texttt{question\_classifier}.},label=lst:drop-classifier]
{
  "spec_id": "question_classifier",
  "role": "Classifies the reasoning operation required
    by the question",
  "tools": ["web_search"],
  "dependencies": ["passage_analyst"],
  "io_contract": {
    "input_schema": {
      "question": "string",
      "numbers": "list -- from passage_analyst",
      "dates": "list -- from passage_analyst",
      "entities": "list -- from passage_analyst",
      "candidate_spans": "list -- from passage_analyst"
    },
    "output_schema": {
      "operation_type": "string -- one of:
        arithmetic_difference | entity_counting |
        span_extraction | sorting_ranking |
        date_duration | aggregation_summation |
        conditional_multihop | negation_exclusion |
        comparison_boolean | unanswerable",
      "reasoning_plan": "list of strings -- numbered
        step-by-step plan for the Reasoning Engine",
      "required_operands": "list of {operand_ref: string,
        source: 'numbers'|'entities'|'dates'|'spans',
        role: string}",
      "confidence": "float 0.0-1.0"
    },
    "description": "Identifies operation type and produces
      a step-by-step reasoning plan with operand refs."
  },
  "verification_criteria": {
    "behavioral_assertions": [
      "Given 'How many more...', must classify as
       'arithmetic_difference'",
      "Given 'How many players...', must classify as
       'entity_counting'",
      "reasoning_plan must contain at least one step"
    ],
    "forbidden_patterns": [
      "Must not compute the final answer",
      "Must not produce more than one operation_type"
    ]
  }
}
\end{lstlisting}

\paragraph{Agent 3: Reasoning Engine} (depends on \texttt{passage\_analyst}, \texttt{question\_classifier}):

\begin{lstlisting}[language={},basicstyle=\ttfamily\scriptsize,breaklines=true,frame=single,caption={Agent spec: \texttt{reasoning\_engine}.},label=lst:drop-engine]
{
  "spec_id": "reasoning_engine",
  "role": "Executes the discrete reasoning operation and
    computes the raw answer",
  "tools": ["file_generator"],
  "dependencies": ["passage_analyst",
                    "question_classifier"],
  "io_contract": {
    "input_schema": {
      "question": "string",
      "numbers": "list -- from passage_analyst",
      "dates": "list -- from passage_analyst",
      "entities": "list -- from passage_analyst",
      "candidate_spans": "list -- from passage_analyst",
      "passage_sentences": "list -- from passage_analyst",
      "operation_type": "string -- from classifier",
      "reasoning_plan": "list -- from classifier",
      "required_operands": "list -- from classifier"
    },
    "output_schema": {
      "raw_answer": "string|number|date",
      "answer_type": "string -- number|span|date",
      "chain_of_thought": "list of strings -- one entry
        per reasoning step with intermediate results",
      "supporting_evidence": "list of strings --
        passage sentences used as evidence"
    },
    "description": "Executes arithmetic, counting,
      sorting, date arithmetic, span extraction, or
      multi-hop reasoning step by step."
  },
  "verification_criteria": {
    "behavioral_assertions": [
      "Given operation_type='arithmetic_difference' with
       operands 14 and 7, raw_answer must be 7",
      "chain_of_thought must contain at least as many
       entries as reasoning_plan steps",
      "supporting_evidence must be non-empty for
       answerable questions"
    ],
    "forbidden_patterns": [
      "Must not introduce facts not present in the
       structured fact table",
      "Must not skip steps listed in reasoning_plan"
    ]
  }
}
\end{lstlisting}

\paragraph{Agent 4: Answer Formatter} (depends on \texttt{reasoning\_engine}, \texttt{question\_classifier}):

\begin{lstlisting}[language={},basicstyle=\ttfamily\scriptsize,breaklines=true,frame=single,caption={Agent spec: \texttt{answer\_formatter}.},label=lst:drop-formatter]
{
  "spec_id": "answer_formatter",
  "role": "Validates, formats, and emits the final
    compliant answer",
  "tools": ["file_generator"],
  "dependencies": ["reasoning_engine",
                    "question_classifier"],
  "io_contract": {
    "input_schema": {
      "question": "string",
      "raw_answer": "string|number|date",
      "answer_type": "string",
      "operation_type": "string",
      "chain_of_thought": "list of strings",
      "supporting_evidence": "list of strings"
    },
    "output_schema": {
      "final_output": "string -- full response with
        chain-of-thought followed by blank line then
        'ANSWER: <formatted_answer>' on the last line",
      "formatted_answer": "string -- the bare answer
        (number, short phrase, or date)",
      "validation_status": "string -- VALID | INVALID
        | UNANSWERABLE"
    },
    "description": "Validates raw_answer against
      answer_type, formats it (strips units, normalizes
      dates, rounds floats), and assembles the final
      output with chain-of-thought followed by
      'ANSWER: <value>' on the last line."
  },
  "verification_criteria": {
    "behavioral_assertions": [
      "final_output must end with a line matching
       'ANSWER: <value>' exactly",
      "formatted_answer must be a bare number, short
       phrase, or date -- not a full sentence",
      "If validation_status is INVALID, final_output
       must still contain the best-supported answer"
    ],
    "forbidden_patterns": [
      "Must not omit the 'ANSWER: ' prefix",
      "Must not emit a full sentence as the answer"
    ]
  }
}
\end{lstlisting}

\subsection{Stage 3: Directive-Driven API Research}

\begin{lstlisting}[language={},basicstyle=\ttfamily\scriptsize,breaklines=true,frame=single,caption={Stage 3: API research results (abbreviated).},label=lst:drop-stage3]
{
  "directive_results": [
    {
      "directive_id": "passage_analyst",
      "research_summary": "Searched for named entity
        recognition and number extraction approaches
        for factual passages..."
    },
    {
      "directive_id": "question_classifier",
      "research_summary": "Searched for discrete reasoning
        operation taxonomies and question classification
        techniques for reading comprehension..."
    },
    {
      "directive_id": "reasoning_engine",
      "research_summary": "Searched for numerical reasoning
        approaches over unstructured text, including
        counting, arithmetic, and date computation..."
    },
    {
      "directive_id": "answer_formatter",
      "research_summary": "Searched for answer formatting
        conventions and validation approaches for
        short-answer reading comprehension..."
    }
  ]
}
\end{lstlisting}

\noindent Duration: 706\,s. Four search directives executed, one per agent spec.

\subsection{Execution Example}

At runtime, the coordinator executes the four agents in topological order. For the problem:

\begin{lstlisting}[language={},basicstyle=\ttfamily\small,breaklines=true,frame=single,caption={Input passage and question.}]
Passage: "Hoping to rebound from their loss to the
Patriots, the Raiders stayed at home for a Week 16 duel
with the Houston Texans. Oakland took the early lead when
Chaz Schilens hauled in a 10-yard touchdown pass from
JaMarcus Russell in the first quarter. The Texans responded
with a 28-yard field goal by Kris Brown. In the second
quarter, Oakland extended its lead with a Sebastian
Janikowski 40-yard field goal."

Question: "How many field goals were kicked?"
\end{lstlisting}

\begin{enumerate}
\item \textbf{Passage Analyst} extracts the structured fact table:
\begin{lstlisting}[language={},basicstyle=\ttfamily\scriptsize,breaklines=true,frame=single]
{
  "numbers": [
    {"value": 10, "unit": "yards",
     "context": "10-yard touchdown pass",
     "entity_ref": "Chaz Schilens"},
    {"value": 28, "unit": "yards",
     "context": "28-yard field goal",
     "entity_ref": "Kris Brown"},
    {"value": 40, "unit": "yards",
     "context": "40-yard field goal",
     "entity_ref": "Sebastian Janikowski"}
  ],
  "entities": [
    {"name": "Chaz Schilens", "type": "person",
     "attributes": [{"key": "play", "value": "TD catch"}]},
    {"name": "Kris Brown", "type": "person",
     "attributes": [{"key": "play", "value": "FG"}]},
    {"name": "Sebastian Janikowski", "type": "person",
     "attributes": [{"key": "play", "value": "FG"}]}
  ]
}
\end{lstlisting}

\item \textbf{Question Classifier} identifies:
\begin{lstlisting}[language={},basicstyle=\ttfamily\scriptsize,breaklines=true,frame=single]
{
  "operation_type": "entity_counting",
  "reasoning_plan": [
    "1. Identify all field goal events in the passage",
    "2. Count the distinct field goal events",
    "3. Return the count"
  ],
  "required_operands": [
    {"operand_ref": "Kris Brown FG",
     "source": "entities", "role": "count_target"},
    {"operand_ref": "Janikowski FG",
     "source": "entities", "role": "count_target"}
  ],
  "confidence": 0.95
}
\end{lstlisting}

\item \textbf{Reasoning Engine} executes:
\begin{lstlisting}[language={},basicstyle=\ttfamily\scriptsize,breaklines=true,frame=single]
{
  "raw_answer": 2,
  "answer_type": "number",
  "chain_of_thought": [
    "Step 1: Field goals in passage: (a) 28-yard FG
     by Kris Brown, (b) 40-yard FG by Janikowski.",
    "Step 2: Count = 2.",
    "Step 3: Verify -- no other field goals mentioned."
  ],
  "supporting_evidence": [
    "The Texans responded with a 28-yard field goal
     by Kris Brown.",
    "Oakland extended its lead with a Sebastian
     Janikowski 40-yard field goal."
  ]
}
\end{lstlisting}

\item \textbf{Answer Formatter} outputs:
\begin{lstlisting}[language={},basicstyle=\ttfamily\scriptsize,breaklines=true,frame=single]
{
  "final_output": "Step 1: Field goals in passage:
    (a) 28-yard FG by Kris Brown, (b) 40-yard FG
    by Janikowski.\nStep 2: Count = 2.\nStep 3:
    Verified -- no other FGs mentioned.\n\nANSWER: 2",
  "formatted_answer": "2",
  "validation_status": "VALID"
}
\end{lstlisting}
\end{enumerate}



\end{document}